\documentclass{bioinfo}

\usepackage{graphicx}
\usepackage{threeparttable}
\usepackage{diagbox}
\usepackage{multirow}
\usepackage{arydshln}

\copyrightyear{2019} \pubyear{2019}

\access{Advance Access Publication Date: Day Month Year}
\appnotes{Original Paper}

\begin{document}
\firstpage{1}

\subtitle{Data and text mining}

\title[CPI Extraction via Gaussian Probability Distribution and External Biomedical Knowledge]{Chemical-protein Interaction Extraction via Gaussian Probability Distribution and External Biomedical Knowledge}
\author[Sun \textit{et~al}.]{Cong Sun\,$^{\text{\sfb 1}}$, Zhihao Yang\,$^{\text{\sfb 1,}*}$, Leilei Su\,$^{\text{\sfb 2}}$, Lei Wang\,$^{\text{\sfb 3,}*}$, Yin Zhang\,$^{\text{\sfb 3}}$, \\Hongfei Lin\,$^{\text{\sfb 1}}$ and Jian Wang\,$^{\text{\sfb 1}}$}
\address{$^{\text{\sf 1}}$School of Computer Science and Technology, Dalian University of Technology, Dalian, 116024, China\\
$^{\text{\sf 2}}$School of Mathematical Sciences, Dalian University of Technology, Dalian, 116024, China\\
$^{\text{\sf 3}}$Beijing Institute of Health Administration and Medical Information, Beijing, 100850, China\\}

\corresp{$^\ast$To whom correspondence should be addressed.}

\history{Received on XXXXX; revised on XXXXX; accepted on XXXXX}

\editor{Associate Editor: XXXXXXX}

\abstract{\textbf{Motivation:} The biomedical literature contains a wealth of chemical-protein interactions (CPIs). Automatically extracting CPIs described in biomedical literature is essential for drug discovery, precision medicine, as well as basic biomedical research. Most existing methods focus only on the sentence sequence to identify these CPIs. However, the local structure of sentences and external biomedical knowledge also contain valuable information. Effective use of such information may improve the performance of CPI extraction.\\
\textbf{Results:} In this paper, we propose a novel neural network-based approach to improve CPI extraction. Specifically, the approach first employs BERT to generate high-quality contextual representations of the title sequence, instance sequence, and knowledge sequence. Then, the Gaussian probability distribution is introduced to capture the local structure of the instance. Meanwhile, the attention mechanism is applied to fuse the title information and biomedical knowledge, respectively. Finally, the related representations are concatenated and fed into the softmax function to extract CPIs. We evaluate our proposed model on the CHEMPROT corpus. Our proposed model is superior in performance as compared with other state-of-the-art models. The experimental results show that the Gaussian probability distribution and external knowledge are complementary to each other. Integrating them can effectively improve the CPI extraction performance. Furthermore, the Gaussian probability distribution can effectively improve the extraction performance of sentences with overlapping relations in biomedical relation extraction tasks.\\
\textbf{Availability:} Data and code are available at https://github.com/CongSun-dlut/CPI\_extraction.\\
\textbf{Contact:} \href{yangzh@dlut.edu.cn, wangleibihami@gmail.com}{yangzh@dlut.edu.cn, wangleibihami@gmail.com}\\
\textbf{Supplementary information:} Supplementary data are available at \textit{Bioinformatics}
online.}

\maketitle

\section{Introduction}
Knowledge of chemical-protein interactions (CPIs) is essential for drug discovery, precision medicine, as well as basic biomedical research \citep{krallinger2017overview}. Currently, PubMed has included approximately 30 million articles and continues to grow at a rate of more than one million articles per year. Extensive valuable CPI knowledge is hidden in biomedical texts, and how to accurately and automatically extract these interactions is increasingly attracting interest. Therefore, the automatic extraction of CPIs from biomedical literature is becoming a crucial task in biomedical natural language processing (NLP) tasks. Recently, the BioCreative VI task organizers released a CHEMPROT corpus, which is the first corpus for CPI extraction \citep{krallinger2017overview}.

\begin{figure*}[tb]
\centerline{\includegraphics[width=1.0\linewidth]{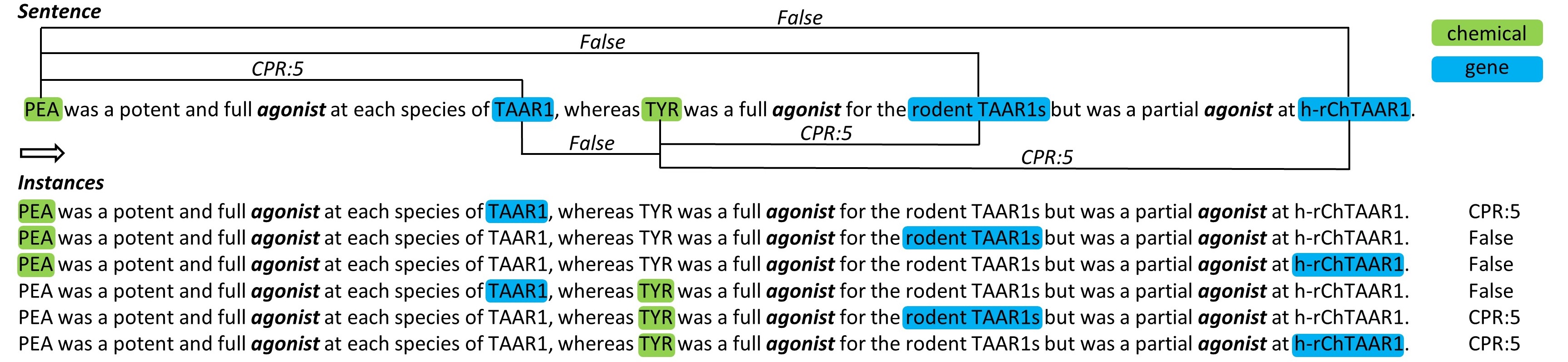}}
\caption{An example sentence which has six overlapping relations. Chemical and protein entities are labeled green and blue, respectively. This example shows how a sentence with overlapping relations generates instances.}
\end{figure*}

Most existing CPI extraction methods can be roughly categorized into two classes: statistical machine learning-based methods and neural network-based methods. Statistical machine learning-based methods usually exploit elaborate kernel functions or explicit feature engineering to extract CPIs. For example, \citep{warikoo2018lptk} utilized a linguistic pattern-aware dependency tree kernel to extract CPIs. \citep{lung2019extracting} constructed chemical-protein relation (CPR) pairs and triplets and exploited sophisticated features to implement CPI extraction. In general, the performance of these methods depends heavily on the designed kernel function and chosen feature set, which is an empirical and skill-dependent work. Compared with statistical machine learning-based methods, neural network-based methods can automatically learn latent features and have become a dominant method for CPI extraction. For example, \citep{peng2018extracting} proposed a method integrating a support vector machine (SVM) \citep{cortes1995support}, a convolutional neural network (CNN) \citep{kim2014convolutional} and a recurrent neural network (RNN) to extract CPIs. \citep{sun2019deep} attempted to improve CPI extraction by introducing the entity attention and ELMo representations \citep{peters2018deep} into the bidirectional long short-term memory (Bi-LSTM) \citep{hochreiter1997long}. Despite the success of these neural networks, some disadvantages remain. First, the pre-trained word embeddings \citep{mikolov2013efficient,pennington2014glove,bojanowski2017enriching} can only learn a context-independent representation for each word \citep{peters2018deep}. Second, CNNs and RNNs perform distinctly worse than the self-attention on word sense disambiguation \citep{tang-etal-2018-self}. Recently, \citep{devlin2018bert} proposed a language representation model called BERT, which stands for bidirectional encoder representations from Transformers \citep{vaswani2017attention}. Compared with previous neural network-based models, BERT eschews the disadvantages of CNNs and RNNs and addresses the drawbacks of pre-trained word embeddings. In the biomedical domain, \citep{lee2019biobert} introduced BioBERT (Bidirectional Encoder Representations from Transformers for Biomedical Text Mining) to improve CPI extraction and they achieve a state-of-the-art performance of CPI extraction. \citep{peng-etal-2019-transfer} made a comprehensive comparison of various BERTs and introduced the BLUE (biomedical language understanding evaluation) benchmark to facilitate research. In their experiments, BERT pre-trained on PubMed obtains a state-of-the-art performance of CPI extraction. Overall, neural network-based methods can automatically learn latent features from vast amounts of unlabeled biomedical texts, thereby achieving state-of-the-art performances.

To date, research on CPI extraction is still at an early stage, and the performance has much room to improve. As shown in Figure 1, one sentence with overlapping relations will generate multiple identical instances based on different target entity pairs. For these identical instances, we refer to them as overlapping instances. Correspondingly, we refer to the sentences that have only one pair of target entities as normal instances. Both the above-mentioned methods do not consider the impact of overlapping instances. However, there are a large number of such overlapping instances in the CHEMPROT corpus, and effectively extracting them may improve the overall performance. Even in the NLP field, the relation extraction of sentences with overlapping relations is also a research difficulty. \citep{zeng-etal-2018-extracting} proposed an end-to-end neural model based on the sequence-to-sequence learning framework with a copy mechanism for relational fact extraction. \citep{Takanobu-etal-2019} applied a hierarchical reinforcement learning framework to enhance the interaction between entity mentions and relation types. To the best of our knowledge, only these two works have studied the extraction of overlapping relations by jointly extracting both entity mentions and relation types, and there is currently no work that focuses on the research of overlapping instances in biomedical relation extraction tasks.

Our work focuses on improving the CPI extraction performance of overlapping instances and starts from the following two aspects. On the one hand, it is more reasonable that the relation extraction model should focus on the target entity and its adjacent words than on treating each token in the instance equally. Existing models usually use position features (e.g., relative distance embeddings) to indicate the position of each token in the instance, which can help the model learn different latent information for overlapping instances to some extent. Nevertheless, because the word/token sequence of the instance is the major information for neural network-based methods \citep{zhang2019neural}, it is difficult to effectively improve the extraction performance of overlapping instances by using these position features. Another commonly used method is to use the attention mechanism to calculate the weight of each token at different positions in overlapping instances. Take the first instance in Figure 1 as an example. Given a pre-trained matrix of word embeddings, this instance can be represented as a sequence of word embeddings, where the tokens 'PEA' and 'TAAR1' are two target entities. In general, the attention mechanism can use the embeddings of each token and target entities to calculate the weight, and thus it can distinguish the importance of different tokens. However, for the word 'agonist' in the instance, it appears in three different positions and shares the same word embedding. The attention mechanism can only use the word embedding of 'agonist' and the word embedding of the target entities to calculate the weight of 'agonist'; therefore, it cannot distinguish the importance of the three 'agonist' in the instance. Intuitively, adjacent words usually contribute more semantically to the target entities than distant ones \citep{guo2019gaussian}, and thus the first 'agonist' should be more important to the target entities than the other two. To overcome these limitations of existing methods, we introduce the Gaussian probability distribution to enhance the weights of the target entity and its adjacent words. 

On the other hand, the CHEMPROT corpus is a corpus composed of abstracts from PubMed. The title information and the knowledge from the biomedical knowledge base may have a positive impact on CPI extraction. For example, in the abstract file of the CHEMPROT corpus, the title "Loperamide modifies but does not block the corticotropin-releasing hormone-induced ACTH response in patients with Addison's disease" contains vital information on the entities 'loperamide' and 'ACTH'. This information may be helpful for correctly predicting the interaction between these two entities. Moreover, Therapeutic Target Database (TTD) \citep{li2017therapeutic} is a knowledge base to provide information about the known protein and the corresponding chemicals directed at each target protein. We can use the knowledge from TTD to generate CPR tags for instances. Therefore, we further leverage the title information and biomedical knowledge (collectively referred to as external knowledge) to guide CPI extraction.

In this paper, we propose a novel neural network-based approach to improve the CPI extraction performance of overlapping instances. Specifically, the approach first employs BERT to generate high-quality contextual representations of the title sequence, instance sequence, and knowledge sequence, respectively. Then the Gaussian probability distribution is introduced to enhance the weights of the target entity and its adjacent words in the instance sequence. Meanwhile, the attention mechanism is applied to fuse the title information and biomedical knowledge. Finally, the related representations are concatenated and fed into the softmax function to extract CPIs. The contributions of our study are three-fold.

\begin{itemize}
\item We propose a novel neural network-based approach to improve the CPI extraction performance of overlapping instances. The experimental results show that our proposed model can improve the extraction performance of overlapping instances while maintaining the performance of normal instances.

\item We introduce the Gaussian probability distribution and external knowledge into our proposed model to improve CPI extraction. Through extensive experiments on the CHEMPROT corpus, we demonstrate that the Gaussian probability distribution and external knowledge are helpful for CPI extraction, and they are highly complementary to each other.

\item To the best of our knowledge, this is the first study that introduces the Gaussian probability distribution into biomedical relation extraction tasks. Through extensive experiments on the CHEMPROT and DDIExtraction 2013 corpora, we demonstrate that the Gaussian probability distribution can effectively improve the extraction performance of overlapping instances.

\vspace*{1pt}
\end{itemize}

\section{Methods}
\subsection{CPI Extraction}

\begin{figure*}[tb]
\centerline{\includegraphics[width=1.0\linewidth]{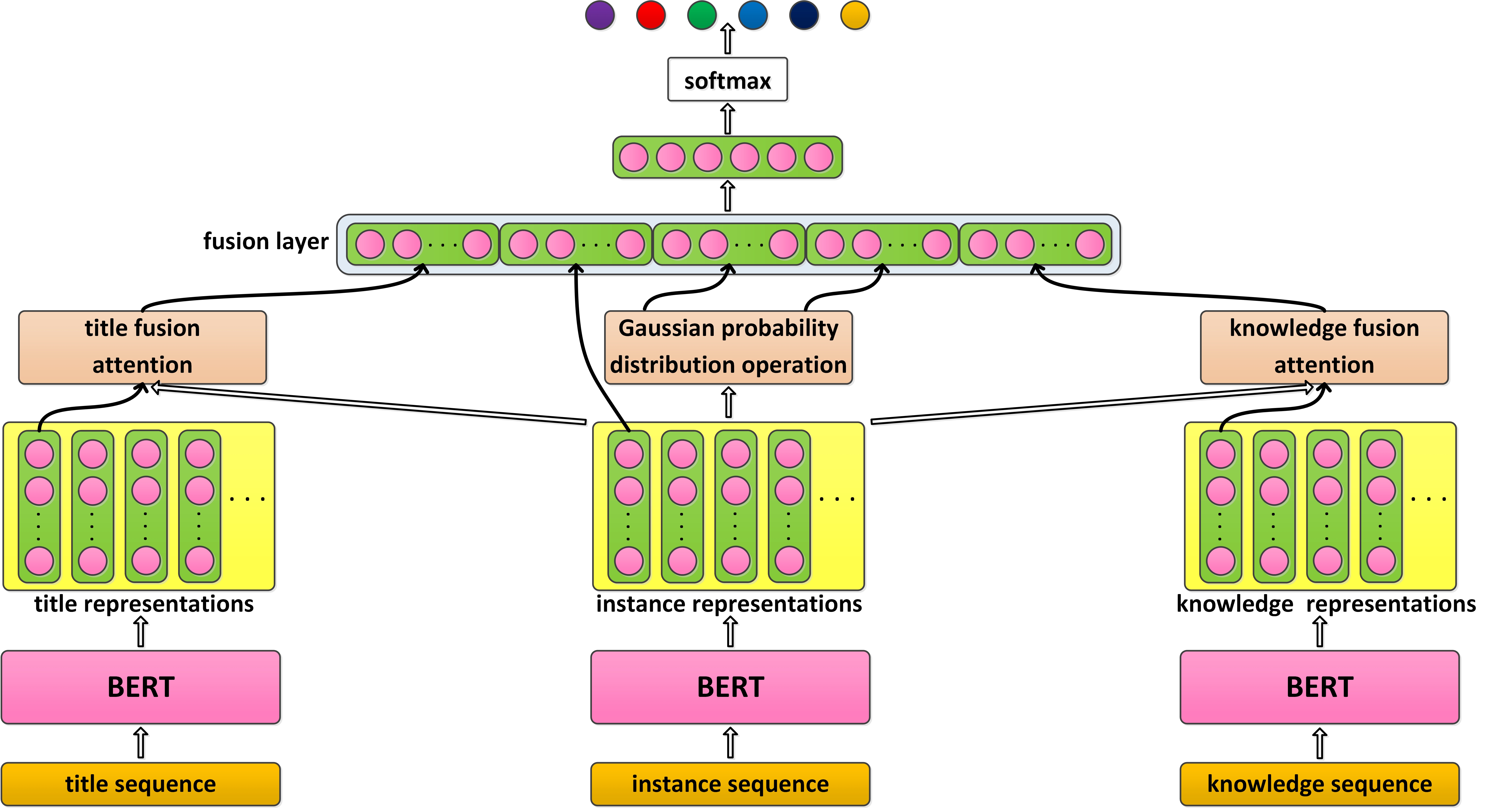}}
\caption{The schematic overview of our proposed model.}
\end{figure*}

CPI extraction is a task for detecting whether a specified CPR type between the target chemical-protein pair within a sentence or document and if so, classifying the CPR type. The CHEMPROT corpus contains five types (i.e., 'CPR:3', 'CPR:4', 'CPR:5', 'CPR:6' and 'CPR:9') used for evaluation purposes. Therefore, we formulate CPI extraction into a multi-class classification problem (including the false relation). The classification problem is defined as follows: given the instances $\{x_1, \cdots, x_i, \cdots, x_n\}$, the goal is to classify the relation $r$ for $x_i$. Essentially, our proposed model is to estimate the probability $P (r|x_i)$, where $R$ = \{CPR:3, CPR:4, CPR:5, CPR:6, CPR:9 and False\}, $ r \in R $, $ 1 \leq i \leq n $.

\subsection{Model Architecture}

Figure 2 is a schematic overview of our proposed model. Overall, our model consists of three parts: BERT representations, Gaussian probability distribution operation and external knowledge acquisition. The input of our model is the title sequence, instance sequence, and knowledge sequence. We first used BERT to generate high-quality contextual representations of these sequences, respectively. Then the Gaussian probability distribution was introduced to enhance the weights of the target entity and its adjacent words in the instance sequence. Meanwhile, the attention mechanism was applied to fuse the title information and biomedical knowledge, respectively. Finally, the related representations were concatenated and fed into the softmax function to extract CPIs. In the following, our proposed model will be described in detail.

\subsubsection{BERT Representations}

BERT is composed of a multi-layer bidirectional Transformer encoder. For the input sequence, BERT uses the WordPiece tokenizer \citep{devlin2018bert} to alleviate the out-of-vocabulary problem to a certain extent. The first token of every sequence is always a special token 'CLS', and the final hidden state corresponding to this token is used as the sequence representations for classification tasks. More details about BERT can be found in the study \citep{devlin2018bert}.

BERT uses the final hidden state of the token 'CLS' to represent the whole sequence representation. However, we argue that the final hidden states of the remaining tokens also contain valuable information. As a result, we retain all final hidden states of the tokens in the sequence. Given an sequence ($S_i$ = \{$w_1, \cdots, w_i, \cdots, w_N$\}) as input, BERT can be formulated as follows:
\begin{equation}
h_i^0 = W_e w_i + W_b
\end{equation}
\begin{equation}
h_i^l = transformer\_block(h_i^{l-1})
\end{equation}
\begin{equation}
t_i^L = h_i^L
\end{equation}
\begin{equation}
R^{seq} = h_{C\!L\!S}^L
\end{equation}
where $w_i$ is the $i$-th token, $L$ is the total number of layers for BERT, $l$ ($ 1 \leq l \leq L $) is the $l$-th layer, and $N$ denotes the sequence length. Equation 1 indicates input embeddings, Equation 3 denotes the representation of the $i$-th token, and Equation 4 denotes the representations of the sequence. The transformer\_block in Equation 2 contains multi-head attention layers, fully connected layers, and the output layer. Furthermore, the parameters $W_e$, $W_b$, and transformer\_block are pre-trained on large-scale corpora using two unsupervised pre-training tasks, masked language model and next sentence prediction. In the experiments, the input of our proposed model is the title sequence, instance sequence and knowledge sequence. We use BERT to generate these representations, where $Q_t$, $Q_k$, $R^{ins}$ and $u_i$ denote the title sequence representations, the knowledge sequence representations, the instance sequence representations and the instance token representations, respectively.

\subsubsection{Gaussian Probability Distribution Operation}

In this study, we introduce the Gaussian probability distribution to enhance the weights of the target entity and its adjacent words, so that the model can learn the local structure of the instance. The Gaussian probability density function is:
\begin{equation}
f(x) = \frac{1}{\sqrt{2\pi}\delta}\exp(-\frac{(x-\mu)^2}{2\delta^2})
\end{equation}
the Gaussian cumulative distribution function is:
\begin{equation}
F(x) = \int_{-\infty}^{x}f(x)\,dx
\end{equation}
the Gaussian probability distribution function is:
\begin{equation}
P(x) = F(x) - F(x-w)
\end{equation}
where $x$ is a real number, $\mu$ is the mean of the distribution, $\delta$ is the standard deviation, and $w$ is the token window. In the experiments, we set the token window $w$ to 1 to represent the distance of each token itself, and set the optimal values of $\mu$ and $\delta$ to 0 and 2.5, respectively. These parameters are tuned on the development set.

\begin{figure}[b]
\centering{\includegraphics[width=1.0\linewidth]{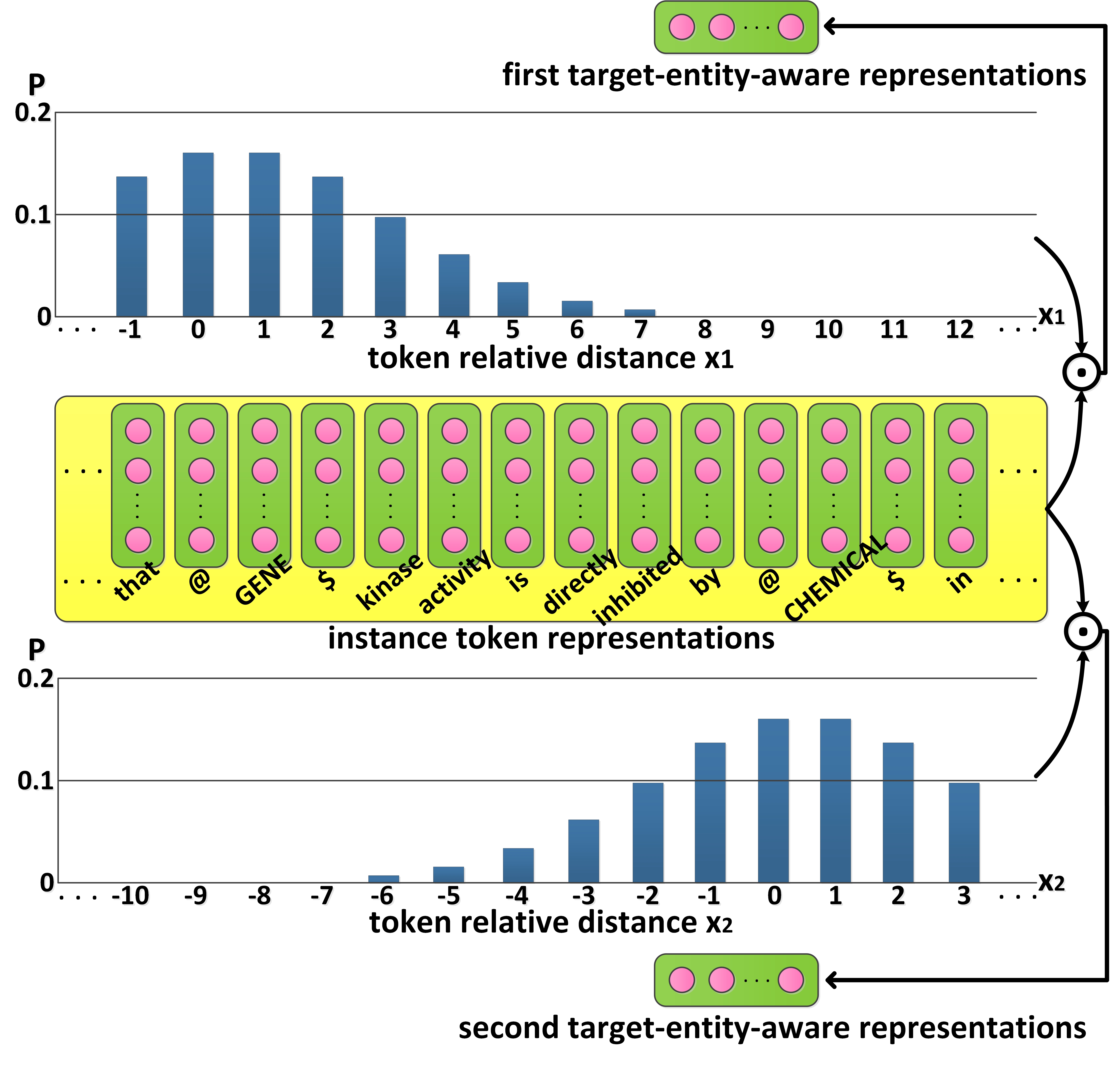}}
\caption{The process of Gaussian probability distribution operation.}
\end{figure}

Figure 3 illustrates the Gaussian probability distribution operation process for tokens in the instance. We first numbered the tokens according to the two target entities in the instance, thus obtaining two token relative distance lists ($x_1$ and $x_2$ in Figure 3). Then we used the Gaussian probability distribution function to calculate the probability of each token in the instance. Afterwards, these token probabilities were performed element-wise multiplication with the token representations. Finally, we obtained the first and second target-entity-aware representations, respectively. The formula is as follows:

\begin{equation}
R^{tar1} = \sum_1^N tanh(P(x_1) u_i)
\end{equation}

\begin{equation}
R^{tar2} = \sum_1^N tanh(P(x_2) u_i)
\end{equation}
where $N$ is the sequence length, $x_1$ and $x_2$ are token numbers, and $R^{tar1}$ and $R^{tar2}$ denote the first and second target-entity-aware representations, respectively.

\subsubsection{External Knowledge Acquisition}

In this study, we used the title information and knowledge sequence to improve the performance of CPI extraction. For the title information acquisition, we first employed BERT to generate high-quality contextual representations of the title sequence. Then we applied the attention mechanism to calculate the query of the title sequence and the key-value pairs of the instance sequence, thereby incorporating the title information into our proposed model. The title fusion attention is defined as follows:

\begin{equation}
t_i = Q_t^T u_i
\end{equation}

\begin{equation}
\alpha_i = \frac{exp(t_i)}{\sum_n exp(t_n)}
\end{equation}

\begin{equation}
R^{title} = \sum_1^N tanh(\alpha_i u_i)
\end{equation}
where $N$ is the sequence length, $Q_t$ is the representations of the title sequence, $u_i$ is the representation of the $i$-th token, and $R^{title}$ denotes the fused title representations.

Figure 4 illustrates the process for obtaining the biomedical knowledge sequence. We first obtained the CPR tags from TTD based on the target chemical and protein entities of the instances, which we denote by $K$. To improve the quality of $K$, we only obtained the tags of the 'CPR:4', 'CPR:5' and 'CPR:6' types and eliminated the ambiguous CPR tags. Then, we extended $K$ by including tokens in the shortest dependency path (SDP) from the target chemical and protein entities. Thus, we used $K$ as the knowledge sequence. More details are provided in Supplementary Material: \emph{Knowledge Sequence Acquisition}. Afterwards, we also employed BERT to generate high-quality contextual representations of the knowledge sequence. Finally, the attention mechanism is applied to calculate the query of $K$ and the key-value pairs of the instance sequence. The knowledge fusion attention is defined as follows:

\begin{equation}
k_i = Q_k^T u_i
\end{equation}

\begin{equation}
\alpha_i = \frac{exp(k_i)}{\sum_n exp(k_n)}
\end{equation}

\begin{equation}
R^{know} = \sum_1^N tanh(\alpha_i u_i)
\end{equation}
where $N$ is the sequence length, $Q_k$ is the representations of the knowledge sequence, $u_i$ is the representation of the $i$-th token, and $R^{know}$ denotes the fused knowledge representations.

\begin{figure}[htb]
\centerline{\includegraphics[width=1.0\linewidth]{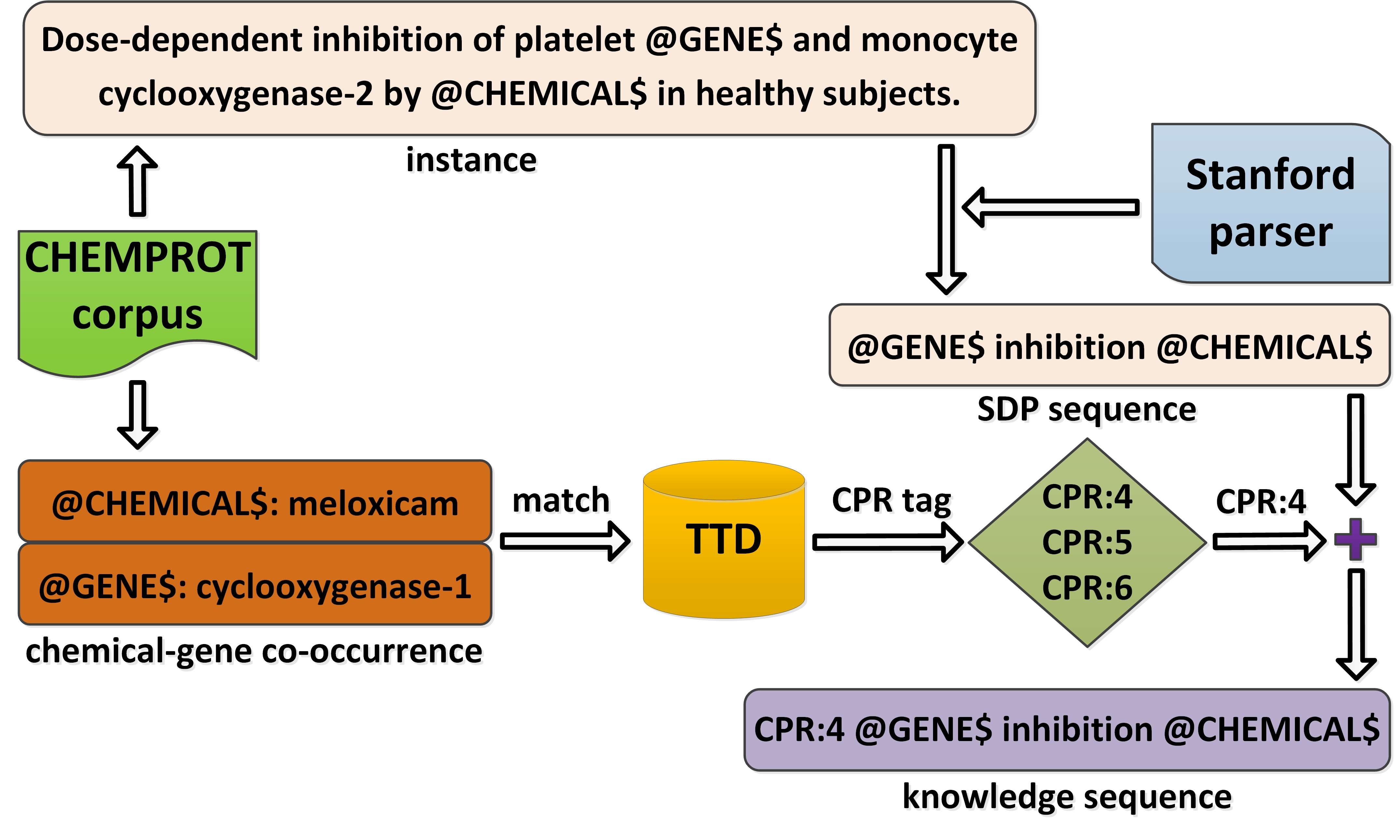}}
\caption{The process for obtaining the biomedical knowledge sequence.}
\end{figure}

\subsubsection{Fusion and Classification}

In the fusion layer, the related representations are concatenated as the fusion representations $h$ ($h$ = $[R^{title};R^{ins};R^{tar1};R^{tar2};R^{know}]$). Then a fully connected neural network is employed to learn the representations $h$. Finally, the softmax function is used to calculate the probability $P$ belonging to the CPR type $r$:

\begin{equation}
P(r|h)=softmax(W_o h + b_o)
\end{equation}
where $W_o$ and $b_o$ are weight parameters and bias parameters. Furthermore, our proposed model uses the categorical cross-entropy function as the loss function and utilizes Adam to optimize the parameters.

\section{Experiments and Discussion}

\subsection{Dataset and Experimental Settings}

In this study, to make a fair comparison with other methods, we used the BLUE CHEMPROT dataset provided by \citep{peng-etal-2019-transfer} to evaluate our proposed model. More detailed descriptions of the CHEMPROT dataset are provided in Supplementary Material: \emph{CHEMPROT Dataset}. The only difference between our instance with Peng's is that we replaced the corresponding chemical and gene mentions with '@CHEMPROT\$' and '@GENE\$', respectively. For example, as shown in Figure 1, the fourth instance is processed as "PEA was a potent and full agonist at each species of @\emph{CHEMICAL}\$, whereas @\emph{GENE}\$ was a full agonist for the rodent TAAR1s but was a partial agonist at h-rChTAAR1." in Peng's instance, but the instance is more reasonable to be processed into our instance "PEA was a potent and full agonist at each species of @\emph{GENE}\$, whereas @\emph{CHEMICAL}\$ was a full agonist for the rodent TAAR1s but was a partial agonist at h-rChTAAR1.". The statistics of evaluated CPR types in the CHEMPROT dataset are illustrated in Table 1.

\begin{table}[htb]
\processtable{Statistics of evaluated CPR types in the CHEMPROT dataset.\label{Tab:01}} {\begin{tabular}{@{}p{1.9cm}p{0.8cm}p{0.8cm}p{0.8cm}p{0.8cm}p{0.8cm}p{0.8cm}@{}}\toprule Set & CPR:3 & CPR:4 & CPR:5 & CPR:6 & CPR:9 & False\\\midrule
Training set &  768 & 2,251  & 173 &235 &727 &15,306 \\
Development set &  550 & 1,094  & 116 &199 &457 &9,404 \\
Test set &  665 & 1,661  & 195 &293 &644  &13,485\\
Total &  1,983 &5,009 & 484 &727 &1,828  &38,195\\
\botrule
\end{tabular}}{}
\end{table}

\begin{table*}[htb]
\processtable{Performance comparison on the CHEMPROT dataset.\label{Tab:02}} {\begin{tabular}{@{}p{3.5cm}p{3.5cm}p{1.0cm}p{1.0cm}p{1.0cm}p{1.0cm}p{1.0cm}p{1.0cm}p{1.0cm}p{1.0cm}@{}}\toprule \multirow{2}{*}{}&
\multirow{2}{*}{Method}&  
\multicolumn{5}{c}{F1 on each CPR type (\%)}&\multicolumn{3}{c}{Overall performance (\%)}\cr  
      \cline{3-10}
&&CPR:3&CPR:4&CPR:5&CPR:6&CPR:9&P&R&F1\cr\midrule
\multirow{2}{*}{Machine learning-based methods}&\citep{warikoo2018lptk} &--&--&--&--&--& 27.32 & 55.14 & 36.54 \\
&\citep{lung2019extracting} &49.80&66.50&56.49&69.64&28.74 &63.52 & 51.21 & 56.71\\
\cdashline{1-10}[2pt/2pt]
\multirow{8}{*}{Neural network-based methods}&\citep{corbett2018improving}\dag &--&--&--&--&--& 56.10 & 67.84 & 61.41 \\
&\citep{mehryary2018potent}\dag &--&--&--&--&-- & 59.05 & 67.76 & 63.10  \\
&\citep{peng2018extracting}\dag &--&--&--&--&-- & 72.66  & 57.35  & 64.10  \\
&\citep{lim2018chemical}\dag&--&--&--&--&-- & 74.8   & 56.0   & 64.1   \\
&\citep{lu2019extracting}  &--&--&--&--&-- & 65.44   &  64.84  & 65.14  \\
&\citep{zhang2019chemical}  &59.4&71.8&65.7&72.5&50.1 & 70.6     &  61.8  & 65.9  \\
&\citep{sun2019deep} &64.69&75.26&68.14&79.26&55.71 & 67.04     &  72.01  & 69.44  \\
&NCBI BERT \citep{peng-etal-2019-transfer}  &--&--&--&--&-- &74.5 &70.6 &72.5\\
&BioBERT \citep{lee2019biobert}  &--&--&--&--&-- &77.02 &75.90 &76.46\\
&Our proposed model &\textbf{71.48}&\textbf{81.28}&\textbf{70.90}&\textbf{79.86}&\textbf{69.87} &\textbf{77.08} & \textbf{76.06} &\textbf{76.56} \\\botrule
\end{tabular}}{Notes. \dag denotes ensemble methods. The '--' denotes the value is not provided in the paper. NCBI BERT and BioBERT refer to the versions that achieve state-of-the-art\\performance on the CHEMPROT dataset, namely NCBI BERT (P) and BioBERTv1.1 (+P). 'P' denotes PubMed. The highest value is shown in bold.}
\end{table*}

In the experiments, we employed the PyTorch (https://pytorch.org/) framework to implement our proposed model. For the BERT model, we used the BERT$_{\!B\!A\!S\!E}$ pre-trained on PubMed provided by \citep{peng-etal-2019-transfer}. The performance of other BERTs was also tested, but no better performance was achieved. The details are provided in Supplementary Material: \emph{Performance on Different BERTs}. We used the training set to train our proposed model and the development set to choose the appropriate hyper-parameters. The test set was only used to evaluate the model. The hyper-parameter settings are consistent throughout this study, and the details are provided in Supplementary Material: \emph{Hyper-parameter Settings}. As most existing methods do, we use the precision (P), recall (R) and F1-score (F1) as metrics. The formula is: $F1$ = 2$PR$$/$($P+R$).

\subsection{Experimental Results}

Table 2 illustrates the experimental results in detail. The first two methods are based on statistical machine learning. These methods exploited a tree kernel function or feature engineering to construct a classifier to implement CPI extraction. However, the performance is not satisfactory. This indicates that it is difficult to effectively achieve CPI extraction using statistical machine learning-based methods. In addition to the first two methods, the others are all neural network-based methods. Among these methods, the first four utilized ensemble methods to obtain better performance. However, the individual model (e.g., CNNs or RNNs) still has much room for improvement. \citep{lu2019extracting} used a granular attention mechanism to enhance the RNN model. \citep{zhang2019chemical} exploited the multi-head attention mechanism and ELMo representations to extract CPIs. \citep{sun2019deep} introduced the entity attention mechanism and ELMo representations to improve performance. These three methods are all single models and obtain competitive performance. Recently, BERT raised the F1-score to a new level. \citep{peng-etal-2019-transfer} and \citep{lee2019biobert} exploited biomedical corpora (e.g., PubMed) to train BERT and used the trained BERT for CPI extraction, respectively. The F1-scores obtained by NCBI BERT \citep{peng-etal-2019-transfer} and BioBERT \citep{lee2019biobert} are 72.5\% and 76.46\%, respectively. Our approach exploited BERT to generate contextual representations and introduced the Gaussian probability distribution and external knowledge to enhance the extraction ability. Our proposed model obtains an F1-score of 76.56\%, which is currently the best performance. These experimental results demonstrate that our proposed model can effectively improve the performance of CPI extraction. Moreover, from Table 2, the performance of BERT and BERT-based models is greatly superior to other neural models. Therefore, we re-implemented both NCBI BERT and BioBERT as baselines to compare with our proposed model and exploited T-TEST to perform statistical significance tests. We employed eight different random seeds to repeat the same experiment eight times under the same experimental settings. The experimental results show that the performance of our proposed model is significantly better than NCBI BERT and BioBERT ($p$ \textless 5\%). In addition, we also compared the performance of each evaluated CPR type compared with other models. Our proposed model obtains the highest F1-score on each CPR type, and more encouragingly, the performance on the extraction of the 'CPR:9' type is greatly superior to other models. This experimental result demonstrates the effectiveness of our proposed model from another aspect.

In conclusion, the experimental results show that our proposed model is superior in performance as compared with other state-of-the-art models. Due to space limitations, the error analysis of our proposed model is provided in Supplementary Material: \emph{Error Analysis}.

\subsection{Extraction of Overlapping and Normal Instances}

\begin{table}[b]
\processtable{Statistics for the instances in the CHEMPROT dataset.\label{Tab:03}} {\begin{tabular}{@{}p{2.55cm}p{1.6cm}p{1.6cm}p{1.6cm}@{}}\toprule  Set & Overlapping & Normal & All\\\midrule
Training set &  18,545 & 915  & 19,460  \\
Development set &  11,303 & 517   &11,820 \\
Test set &  16,132 & 811  &16,943\\
Total &  45,980 & 2,243  &48,223\\
\botrule
\end{tabular}}{}
\end{table}

In this section, we explored the impact of our proposed model on overlapping instances and normal instances. Table 3 shows the statistics for the instances in the CHEMPROT dataset. We first used overlapping instances and normal instances in the test set as input, and then compared the performance of BERT with the performance of our proposed model (the input of our proposed model also contains the title sequence and knowledge sequence). As shown in Table 4, BERT obtains F1-scores of 73.19\% and 83.28\% in the extraction of overlapping instances and normal instances, respectively. Correspondingly, our proposed model achieves F1-scores of 75.89\% and 83.82\%, respectively. These experimental results show that our proposed model can improve the CPI performance of overlapping instances while maintaining the performance of normal instances.

\begin{table}[htb]
\processtable{Overlapping and normal instance extraction on the test set.\label{Tab:04}} {\begin{tabular}{@{}p{2.45cm}p{1.65cm}p{1.0cm}p{1.0cm}p{1.0cm}@{}}\toprule Model &Instances   & P(\%)   & R(\%)  & F1(\%) \\\midrule
BERT (local)   
&overlapping & 73.67  & 72.72   & 73.19  \\
&normal  &81.61  & 85.02   & 83.28  \\
&all & 74.37  & 73.74   & 74.05  \\
Our proposed model   
&overlapping & 76.61  & 75.18   & 75.89  \\
&normal & 82.00  & 85.71   & 83.82  \\ 
&all &77.08 & 76.06 &76.56  \\
\botrule
\end{tabular}}{Notes.'local' denotes the BERT runs locally.}
\end{table}

\subsection{Ablation Study}

In this section, to explore the contribution of each component to overall performance, we performed an ablation study over our proposed model. The experimental results are presented in Table 5. We first explored the impact brought by the Gaussian probability distribution. It can be observed that the 'BERT+Gaussian' model increases the F1-score by 1.44\% compared with BERT. Meanwhile, when the Gaussian probability distribution is removed from our proposed model, the F1-score drops 1.46\%. These experimental results demonstrate that the Gaussian probability distribution is critical in obtaining state-of-the-art performance. Next, we evaluated the impact of external knowledge (i.e., the title information and biomedical knowledge). When the title information or knowledge sequence is added, the F1-score increases by 0.38\% and 1.03\% compared with BERT, respectively. On the other hand, when the title information or knowledge sequence is removed from our proposed model, the F1-score decreases by 1.46\% and 0.99\%, respectively. These experimental results indicate that both the title information and knowledge sequence play an important role in boosting the performance of CPI extraction. Finally, we validated the effects of Gaussian probability distribution and external knowledge. When both removing the Gaussian probability distribution and external knowledge, our proposed model degenerates into the BERT model, and the performance drops to 74.05\% in F1-score. After adding the Gaussian probability distribution or external knowledge, the F1-scores are greatly improved, reaching 75.49\% and 75.10\%, respectively. Moreover, when the Gaussian probability distribution and external knowledge are both added, the F1-score increases to 76.56\%. These experimental results demonstrate that the Gaussian probability distribution and external knowledge are helpful for CPI extraction, and they are highly complementary to each other.

\begin{table}[htb]
\processtable{Ablation study over our proposed model.\label{Tab:05}} {\begin{tabular}{@{}p{3.7cm}p{1.2cm}p{1.2cm}p{1.2cm}@{}}\toprule model   & P(\%)   & R(\%)  & F1(\%) \\\midrule
BERT & 74.37 & 73.74   & 74.05  \\
BERT+Gaussian & 76.69 & 74.32  & 75.49  \\
BERT+title & 74.87 & 74.00   & 74.43  \\
BERT+knowledge & 75.11 & 75.04  & 75.08  \\
BERT+Gaussian+title & \textbf{77.58} & 73.66   & 75.57  \\
BERT+Gaussian+knowledge & 75.98 & 74.26   & 75.11  \\
BERT+title+knowledge & 74.54   & 75.68  & 75.10  \\
Our proposed model (ALL) &77.08 & \textbf{76.06} &\textbf{76.56}\\
\botrule
\end{tabular}}{Notes.'+'denotes adding the corresponding component. 'ALL' denotes\\containing all the components (i.e., BERT+Gaussian+title+knowledge).\\The highest value is shown in bold.}
\end{table}

\subsection{Impact of Gaussian Probability Distribution on Biomedical Relation Extraction}

In this study, we introduced the Gaussian probability distribution into biomedical relation extraction tasks. When removing the external knowledge, our proposed model degenerates into the 'BERT+Gaussian' model. To explore the impact of Gaussian probability distribution on biomedical relation extraction tasks, we further evaluated the 'BERT+Gaussian' model on the CHEMPROT and DDIExtraction 2013 \citep{herrero2013ddi,segura-bedmar-etal-2013-semeval} corpora. DDIExtraction 2013 corpus is a manually annotated drug-drug interaction (DDI) corpus based on the DrugBank database and MEDLINE abstracts. This corpus contains four DDI types for evaluation purposes, namely 'Advice', 'Effect', 'Mechanism' and 'Int'. We also formulate DDI extraction into a multi-class classification problem, and the evaluation metrics of DDI are consistent with CPI, which are precision, recall and F1-score. In the experiments, to fairly compare with existing methods, we also used the BLUE DDIExtraction 2013 dataset provided by \citep{peng-etal-2019-transfer} to evaluate the model. Table 6 illustrates the statistics for the instances in the DDIExtraction 2013 dataset. More detailed descriptions are provided in Supplementary Material: \emph{DDIExtraction 2013 Dataset}.

\begin{table}[htb]
\processtable{Statistics for the instances in the DDIExtraction 2013 dataset.\label{Tab:6}} {\begin{tabular}{@{}p{2.6cm}p{1.6cm}p{1.6cm}p{1.6cm}@{}}\toprule  Set & Overlapping & Normal & All\\\midrule
Training set &  17,584  & 1,195   & 18,779  \\
Development set &  6,850 & 394   &7,244 \\
Test set &  5,426  & 335  &5,761\\
Total &  29,860 & 1,924  &31,784\\
\botrule
\end{tabular}}{}
\end{table}

Table 7 shows the performance comparison on the DDIExtraction 2013 dataset. \citep{zhang2017drug} proposed a hierarchical RNN-based method to integrate the SDP and sentence sequence for DDI extraction. To the best of our knowledge, their method achieved state-of-the-art performance with an F1-score of 72.9\% before BERT appears. \citep{peng-etal-2019-transfer} exploited BERT pre-trained on PubMed to extract DDIs and raised the F1-score to 78.1\%. This experimental result demonstrates the success of BERT on DDI extraction. We used BERT pre-trained on PubMed to generate BERT representations, and employed the 'BERT+Gaussian' model to extract DDIs. The 'BERT+Gaussian' model obtains an F1-score of 82.04\%, which is 3.94\% higher than the corresponding BERT model. This experimental result demonstrates the effectiveness of the Gaussian probability distribution for DDI extraction.

\begin{table}[t]
\processtable{Performance comparison on the DDIExtraction 2013 dataset.\label{Tab:7}} {\begin{tabular}{@{}P{3.35cm}P{1.35cm}P{1.35cm}P{1.35cm}@{}}\toprule Method  & P(\%)   & R(\%)  & F1(\%) \\\midrule
\citep{zhang2017drug} & 74.1 & 71.8 & 72.9 \\
BERT \citep{peng-etal-2019-transfer} &81.1 &75.3 &78.1 \\
BERT+Gaussian &\textbf{83.42} &\textbf{80.69} &\textbf{82.04}\\
\botrule
\end{tabular}}{The highest value is shown in bold.}
\end{table}

\begin{table}[b]
\processtable{The impact of Gaussian probability distribution on biomedical relation extraction.\label{Tab:8}} {\begin{tabular}{@{}P{2.45cm}P{0.6cm}P{1.4cm}P{0.85cm}P{0.85cm}P{0.85cm}@{}}\toprule Model &Task &Instances   & P(\%)   & R(\%)  & F1(\%) \\\midrule
BERT (local) &CPI  
&overlapping & 73.67  & 72.72   & 73.19 \\
&&normal  &81.61  & 85.02   & 83.28  \\
&&all & 74.37  & 73.74   & 74.05  \\
BERT+position &CPI  
&overlapping &73.63  & 73.95   & 73.79 \\
&&normal  &80.46  & 86.06   & 83.16  \\
&&all & 74.23  & 74.96   & 74.59  \\
BERT+entity\_attention &CPI  
&overlapping & 75.26  & 72.82  & 74.02 \\
&&normal  &80.40  & 84.32   & 82.31  \\
&&all & 75.72 & 73.77 & 74.73  \\
BERT+Gaussian & CPI
&overlapping & 76.20  & 73.32   & 74.73  \\
&&normal & 81.67  & 85.37   & 83.48  \\ 
&&all & 76.69  & 74.32   & 75.49  \\
\cdashline{1-6}[2pt/2pt]
BERT (local) & DDI
&overlapping & 79.97  & 76.28   & 78.08  \\
&&normal & 85.87  & 88.76   & 87.29  \\
&&all & 81.12  & 78.55   & 79.81  \\
BERT+position & DDI
&overlapping & 79.58  & 76.40   & 77.96  \\
&&normal & 86.56  & 90.45   & 88.46  \\
&&all & 80.94 & 78.96 & 79.94  \\
BERT+entity\_attention & DDI
&overlapping & 80.88  & 78.15   & 79.49  \\
&&normal & 87.50  & 90.45   & 88.95  \\
&&all & 82.15  & 80.39   & 81.26  \\
BERT+Gaussian & DDI
&overlapping & 82.98   & 79.15 & 81.02   \\
&&normal & 85.25  & 87.64   & 86.43  \\ 
&&all &83.42 &80.69 &82.04  \\
\botrule
\end{tabular}}{Notes.'local' denotes the BERT runs locally.}
\end{table}

Moreover, we further explored the impact of Gaussian probability distribution on biomedical relation extraction. We first used the overlapping instances and normal instances in the CHEMPROT and DDIExtraction 2013 test set as input, respectively. Then, we re-implemented the models of 'BERT+position' and 'BERT+entity\_attention' with reference to Zhang’s work \citep{zhang2017drug}, and compared their performance with the 'BERT+Gaussian' model. As shown in Table 8, compared with BERT, the performance improvement of the 'BERT+position' model on overlapping instances is limited (with +0.6\% and -0.12\% in F1 on CPI and DDI extraction tasks, respectively). This experimental result validates the viewpoint discussed in section \emph{Introduction} that it is difficult to use position features to effectively improve the extraction of overlapping instances \citep{zhang2019neural}. For the 'BERT+entity\_attention' model, its extraction performance on overlapping instances is superior to BERT (with +0.83\% and +1.41\% in F1 on CPI and DDI extraction tasks, respectively) and the 'BERT+position' model (with +0.23\% and +1.53\% in F1 on CPI and DDI extraction tasks, respectively), but inferior to the 'BERT+Gaussian' model (with -0.71\% and -1.53\% in F1 on CPI and DDI extraction tasks, respectively). This experimental result shows that the Gaussian probability distribution plays an important role in boosting the extraction of overlapping instances, and the effect of Gaussian probability distribution is superior to the position features and entity attention mechanism. Through these experiments, we can infer that the Gaussian probability distribution can effectively improve the performance of overlapping instances in biomedical relation extraction tasks. We also list some examples of using the Gaussian probability distribution to correct BERT's predictions. The details are provided in Supplementary Material: \emph{Representative Examples}.

\section{Conclusion}

In this paper, we propose a novel neural network-based approach to improve the performance of CPI extraction by introducing the Gaussian probability distribution and external knowledge. We evaluate our proposed model on the CHEMPROT corpus. It is encouraging to see that the performance of our proposed model outperforms other state-of-the-art models, reaching an F1-score of 76.56\%. The experimental results show that the Gaussian probability distribution and external knowledge are highly complementary to each other. Integrating them can effectively improve the CPI extraction performance. Furthermore, the Gaussian probability distribution can effectively improve the extraction performance of overlapping instances in biomedical relation extraction tasks.

Although our proposed approach exhibits promising performance for CPI extraction from biomedical literature, there is still some room to improve. In future work, we would like to explore the effectiveness of the semi-supervised or unsupervised approach in CPI extraction.

\section*{Funding}

This work was supported by the National Key Research and Development Program of China under Grant 2016YFC0901902.
\\
\\
\emph{Conflict of Interest}: none declared.

\end{document}